\documentclass[manuscript,screen,nonacm]{acmart}

\usepackage{arydshln}
\usepackage{amsmath}
\usepackage{multirow}
\usepackage{xcolor,colortbl}
\newcommand{\ricr}[1]{\textcolor{black}{#1}}
\usepackage{microtype}
\usepackage{graphicx}
\usepackage{subfigure}
\usepackage{nicematrix}
\usepackage{bm}
\usepackage{svg}

\usepackage{algorithm,algpseudocode}

 \newcolumntype{M}[1]{>{\centering\arraybackslash}m{#1}}
 \newcolumntype{?}{!{\vrule width 1pt}}

\AtBeginDocument{%
  \providecommand\BibTeX{{%
    \normalfont B\kern-0.5em{\scshape i\kern-0.25em b}\kern-0.8em\TeX}}}

\setcopyright{acmcopyright}
\copyrightyear{2018}
\acmYear{2018}
\acmDOI{XXXXXXX.XXXXXXX}

\acmConference[Conference acronym 'XX]{Make sure to enter the correct
  conference title from your rights confirmation emai}{June 03--05,
  2018}{Woodstock, NY}
\acmBooktitle{Woodstock '18: ACM Symposium on Neural Gaze Detection,
 June 03--05, 2018, Woodstock, NY} 
\acmPrice{15.00}
\acmISBN{978-1-4503-XXXX-X/18/06}

\begin{document}

\title{CVTT: Cross-Validation Through Time}



\author{Mikhail Andronov}
\authornote{The work was done during an internship in the Tinkoff Lab.}
\email{ext.mandronov@tinkoff.ai}
\affiliation{%
  \institution{Tinkoff}
  \country{}
}

\author{Sergey Kolesnikov}
\email{s.s.kolesnikov@tinkoff.ai}
\affiliation{%
  \institution{Tinkoff}
  \country{}
}
\begin{abstract}
The evaluation of recommender systems from a practical perspective is a topic of ongoing discourse within the research community. While many current evaluation methods reduce performance to a single value metric as an easy way to compare models, it relies on the assumption that the methods' performance remains constant over time. In this study, we examine this assumption and propose the Cross-Validation Thought Time (CVTT) technique as a more comprehensive evaluation method, focusing on model performance over time. By utilizing the proposed technique, we conduct an in-depth analysis of the performance of popular RecSys algorithms. Our findings indicate that (1) the performance of the recommenders varies over time for all reviewed datasets, (2) using simple evaluation approaches can lead to a substantial decrease in performance in real-world evaluation scenarios, and (3) excessive data usage can lead to suboptimal results.

\end{abstract}

\maketitle

\section{Introduction}
Recommender systems (RS) are widely used to personalize the experience of users on various platforms, such as e-commerce websites \cite{bianchiFantasticEmbeddingsHow2020a}, streaming services \cite{bennett2007netflix}, and social media \cite{Wang2021ExploringLT}. The goal of RS is to learn users' preferences and present them with relevant items, such as products \cite{10.1145/3451964.3451966}, songs \cite{10.1145/3240323.3240342}, or videos \cite{10.1145/3289600.3290999}. One of the most critical tasks in RS usage is to measure their offline performance – the performance of RS using historical data, which is usually represented as a single-value result (e.g. NDCG@10 = 0.1). Using a single value measure for RS evaluation can be problematic, as it implicitly assumes RS's consistent performance over time \cite{44265d57d70f4725bf739c67800a5f77}. 
For example, in the case of "crossing" performance over time, one algorithm can consistently increase its performance while another - decrease, resulting in the same single-value \textit{mean} performance. Based on such aggregation, the best algorithm at the evaluation time may not continue to be the best one in the future.

While early works \cite{campos2011towards, timetoconsidertime} have emphasized the importance of time-based validation of algorithms, this practice has not been consistently used in more recent research.
According to prior studies \cite{daisyrec, 10.1145/3383313.3418479}, the majority of RS papers are still utilizing the most straightforward approaches (mostly non-time-dependent) with single-value measurements for evaluation.
On the other hand, several RS papers do report metrics over time \cite{jiCriticalStudyData2022, TTRS}, highlighting the importance of fair data usage for evaluation correctness.

Following the latest RS research trends, in this paper, we present CVTT (\autoref{fig:teaser}) - Cross-validation through time - a simple yet general approach for temporal evaluation of RecSys methods, which follows the most strict and realistic setting for evaluation \cite{10.1145/3383313.3418479} and  supports hyperparameters optimization over time. While validation over time is resource-demanding, such a realistic setting for time-based algorithm validation is precisely what current RecSys evaluation progress insists on \cite{44265d57d70f4725bf739c67800a5f77}.

To summarize, the main contributions of this paper are as follows:
\begin{itemize}
\item We present CVTT, a surprisingly simple method based on recent advances in data splitting and cross-validation techniques. Our method is general and requires minimal changes to be incorporated with most RecSys algorithms. We demonstrate via experiments how commonly used evaluation protocols do not provide adequate modeling of real-world deployment settings compared to CVTT.
\item We show that the performance of RecSys models gradually changes, and our findings demonstrate that new data flow can lead to significant performance changes. 
\item  Finally, we demonstrate that performance can change significantly depending on the chosen data strategy, and continuously extending the training dataset can lead to suboptimal results.
\end{itemize}

\section{Background}

\subsection{Offline evaluation}

Following recent studies in RecSys evaluation \cite{offline_eval_2020, 10.1145/3383313.3418479, DBLP:journals/corr/abs-2010-11060}, we summarize the five main data splitting strategies for offline RecSys evaluation:

\begin{itemize}
\item \textbf{Random Split} randomly splits per-user interactions into train and test folds. The main disadvantage of these methods is that they can hardly be reproduced unless the authors release the used data splits.
\item \textbf{User Split} randomly split users, rather than their interactions, into train and test groups. In this case, particular users and all their interactions are reserved for training, while a different user set and all their interactions are applied for testing. 
\item \textbf{Leave One Out Split} selects one last final transaction per user for testing while keeping all remaining interactions for training. In the case of next-item recommendations, the last interaction corresponds to the last user-item pair per user \cite{he2017neural, bai2019ctrec, zamani2020learning}. In the case of next-basket recommendations, the last interaction is defined as a basket
\cite{rendle2010factorizing, wan2018representing}.
\item \textbf{Temporal User Split} splits per-user interactions into train and test sets based on interaction timestamps (e.g., the last 20\% of interactions are used for testing). While this scenario is actively used in the sequential RecSys domain \cite{liang2018variational, wang2019neural}, it could lead to a data leakage discussed in \cite{10.1145/3383313.3418479}. 
\item \textbf{Temporal Global Split} splits all available interactions into train and test folds based on a fixed timestamp. Interactions before it are used for training, and those that come after are reserved for testing. Compared to Leave One Out Split or Temporal User Split, this method could sample fewer interactions for testing since having the same amount of users or items in train and test sets is not guaranteed. Nevertheless, according to recent studies \cite{DBLP:journals/corr/abs-2010-11060}, this is the only strategy that prevents data leakage.
\end{itemize}

As noted in previous works \cite{10.1145/3383313.3418479} \footnote{We renewed the corresponding data splitting strategies table under Appendix - see \autoref{tab:models}.}, there is little consistency in the selection of evaluation protocols. Even when the same datasets are used, researchers can select different data-splitting strategies for model comparison. For example, in \cite{lu2018coevolutionary} and \cite{yu2019multi} both authors used Amazon and Yelp datasets, but different methods were used as data-splitting strategies: Random Split and Leave One Out, respectively.
On the other hand, a recent study \cite{DBLP:journals/corr/abs-2010-11060} shows that very few (3 out of 30) methods were evaluated in the most realistic and data-leak-proofed strategy by using a global temporal split.
As discussed in Section 3, CVTT focuses on correct RS evaluation over time, thus extending the temporal global split strategy even further.

\subsection{Time-Aware Cross-Validation}

Researchers were interested in time-aware evaluation from as far back as \cite{campos2011towards,Ding_Li_2005}. 
At the same time, while recent works \cite{multiperiodbasket} focus on sequence-aware approaches, they lack simple statistical and matrix factorization-based baselines, benchmarking only RNN-based methods. 
In addition, many recent works \cite{44265d57d70f4725bf739c67800a5f77, sachdeva2019sequential, hu2019sets2sets} only focus on temporal user split evaluation, which may be affected by data leak issue \cite{10.1145/3383313.3418479}. 
Finally, available global temporal-based benchmarks \cite{TTRS} lack time-dependent evaluation over time, reducing the final performance to a single-value metric.
CVTT extends and generalizes recent works \cite{44265d57d70f4725bf739c67800a5f77, 10.1145/3383313.3418479} to address the challenge of Recommender System evaluation over time.

\begin{figure}[t!]
    \centering
    \includegraphics[width=0.9\textwidth]{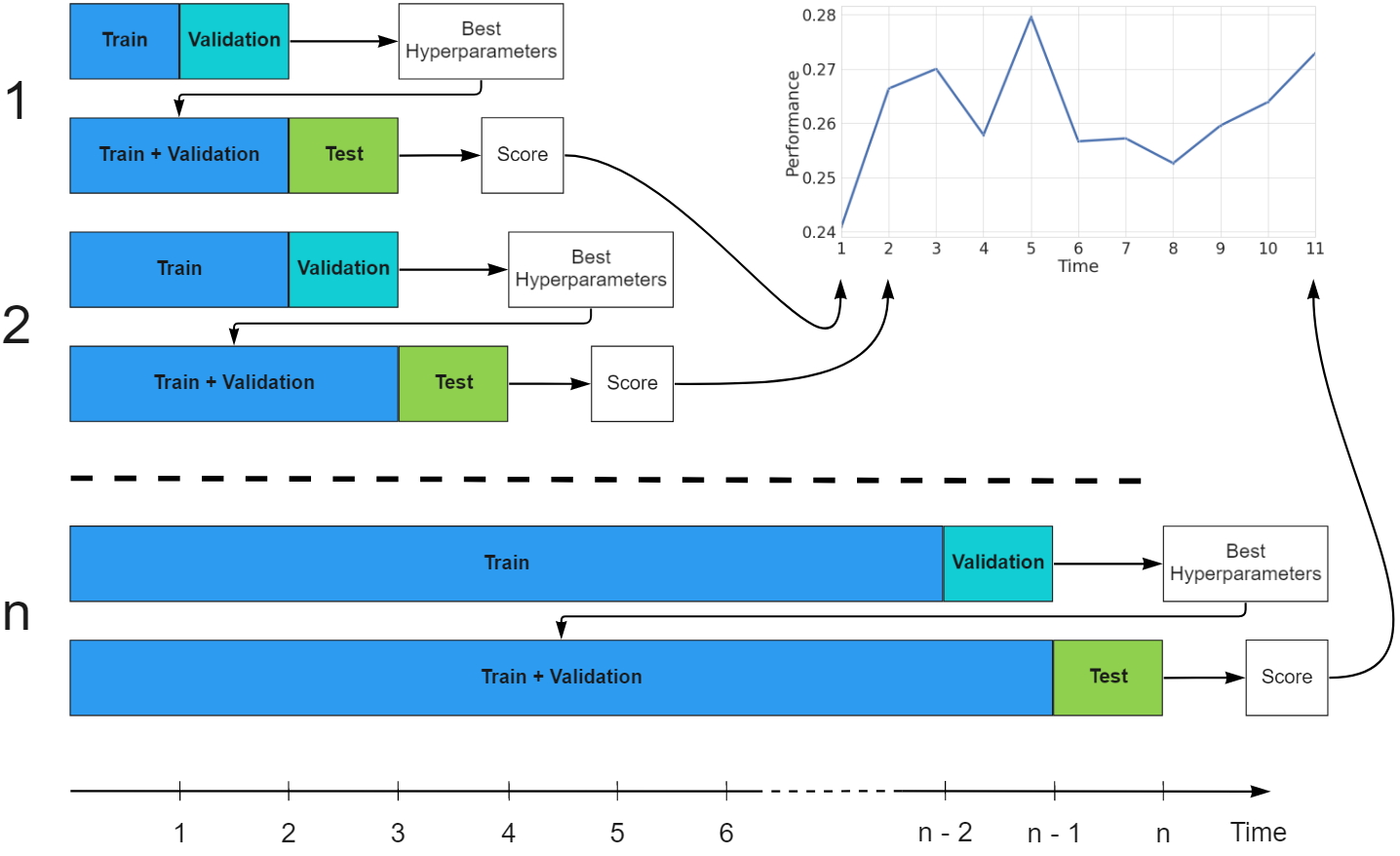}
    \caption{CVTT evaluation strategy scheme. The blue boxes represent data used for training, the turquoise boxes represent data for validation, and the green boxes represent data for testing. At each time step, the following iterations occur: first, training and validation data are used to select the best hyperparameters of the model; then the model with the selected hyperparameters is trained on the combined training and validation data and is evaluated on the test data. In the end, we get the CVTT performance graph by combining these test scores over time.
    }
    \label{fig:teaser}
\end{figure}

\section{Cross-Validation through time}
\label{sec:cvtt}

To evaluate how the recommender's performance changes over time, we adopt a recently proposed global temporal split for cross-validation setup, building on the ideas from \cite{TTRS, DBLP:journals/corr/abs-2010-11060, 44265d57d70f4725bf739c67800a5f77}. 
We split the datasets at regular intervals, leading to several "\textit{folds}" per dataset.
There are two possible ways of creating a training dataset for each fold in this scenario. Each fold could contain all data up to a specific time point (expand strategy) or include only the data from several last intervals (window-$N$ strategy). While we find these two approaches equally fair for real-life applications \footnote{Comparison of the effects of train data strategy selection can be found in Section 5.2.}, we will use the first option (expand) for CVTT overview.

The entire CVTT procedure is shown in \autoref{fig:teaser}.
For each fold with $N$ periods, we split the data into the $"train"$, $"validation"$, and $"test"$ parts using a global temporal split. $"Test"$ (1 period) represents the data from the last period, $"validation"$ (1 period) - penultimate period, and $"train"$ (N-2 periods) - all data up to the penultimate period. Next, we ran a model hyperparameter search on $("train", "validation")$ subset and used the best-found hyperparameters to train and evaluate the model on the $("train+validation", "test")$ one. This $"test"$ score is then used as a final model performance measure on that fold. In the end, we get the CVTT performance graph by combining these scores over time.

\section{Experimental Evaluations}
\label{sec:experiments}

To illustrate the need for CVTT evaluation approach, we tackle the following questions:
(1) How does the CVTT evaluation approach differ from the commonly used random split one? (2) How does model performance vary over time? (3) How do different data preparation strategies affect model performance over time?

\subsection{Datasets}
\label{sec:datasets}

\begin{table*}[!tbp]
\footnotesize
  \centering
    \begin{tabular}{c?c|c|c?c|c|c?c|c|c|c} \toprule
    &\multicolumn{3}{c?}{Before preprocessing}&\multicolumn{3}{c?}{After preprocessing}&\multicolumn{4}{c}{Final statistics}\\ \cmidrule{2-11}
    \multirow{2}{*}{Dataset}&\multirow{3}{*}{\# users}&\multirow{3}{*}{\# items}&\multirow{3}{*}{\# interactions}&\multirow{3}{*}{\# users}&\multirow{3}{*}{\# items}&\multirow{3}{*}{\# interactions}&\# inter.&\# inter. &\# inter. & \# inter. \\ 
      &&&&&&& per user        & per user       & per item       & per item\\         
          &&&&&&&&  per period &        & per period\\  \midrule
    Amazon Movies     & 889k    & 253k     & 7.9m& 6135    & 54k    & 208k & 34 & 6.8 & 3.8 & 1    \\ 
    MovieLens 20M & 138k    & 27k     & 20m &126k   & 26k    & 16m & 129 & 56 & 627 & 15   \\
    Yelp & 1.9m    & 150k     & 6.9m & 8297   & 62k    & 239k & 28 & 5 & 3.8 & 1   \\
    
   	\bottomrule
    \end{tabular}
  \caption{Dataset statistics before and after preprocessing. We measured the number of users, items, and interactions in millions (m) and thousands (k). For the final statistics, we used one month as a period.}

\label{tab:datastat}
\centering

\end{table*}{}

We choose the three most popular time-based datasets in recommender system research:
\begin{itemize}
    \item Amazon Movies and TV - movie recommendation dataset provided by Amazon \cite{li2020time, bai2019ctrec}. It contains over 8 million interactions between almost 900,000 users and more than 250,000 Movies and TV shows from May 1996 to October 2018.
    \item MovieLens 20M - another movie recommendation dataset \cite{xiao2017fairness, li2020time}, that contains 20,000,000 ratings and 465,000 tag applications applied to 27,000 movies by 138,000 users.
    \item Yelp - a business reviews dataset that is well-known among the research community \cite{wang2019kgat, wang2019neural, DBLP:journals/corr/abs-1905-11691}. Contains over 900,000 tips by almost 2,000,000 users.
\end{itemize}
The choice of datasets for our research question was a bit limited, as the datasets need to have timestamps included so that the data can be spit and evaluated over time. 
As a preprocessing step, we also filter the dataset to only include users and items with at least one interaction at each period, following prior works \cite{TTRS}.
Statistical information on the raw datasets is summarized in \autoref{tab:datastat}. 

\begin{figure}[t!]
    \centering
    \begin{subfigure}[]
        \centering
        \includegraphics[width=0.32\textwidth]{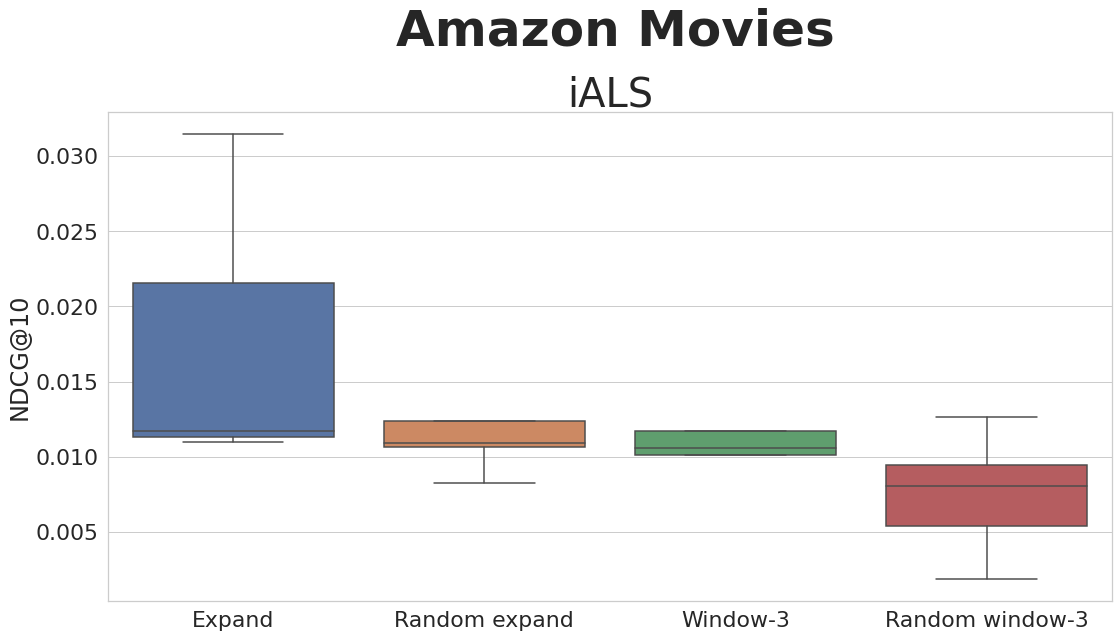}
    \end{subfigure}
    \begin{subfigure}[]
        \centering
        \includegraphics[width=0.32\textwidth]{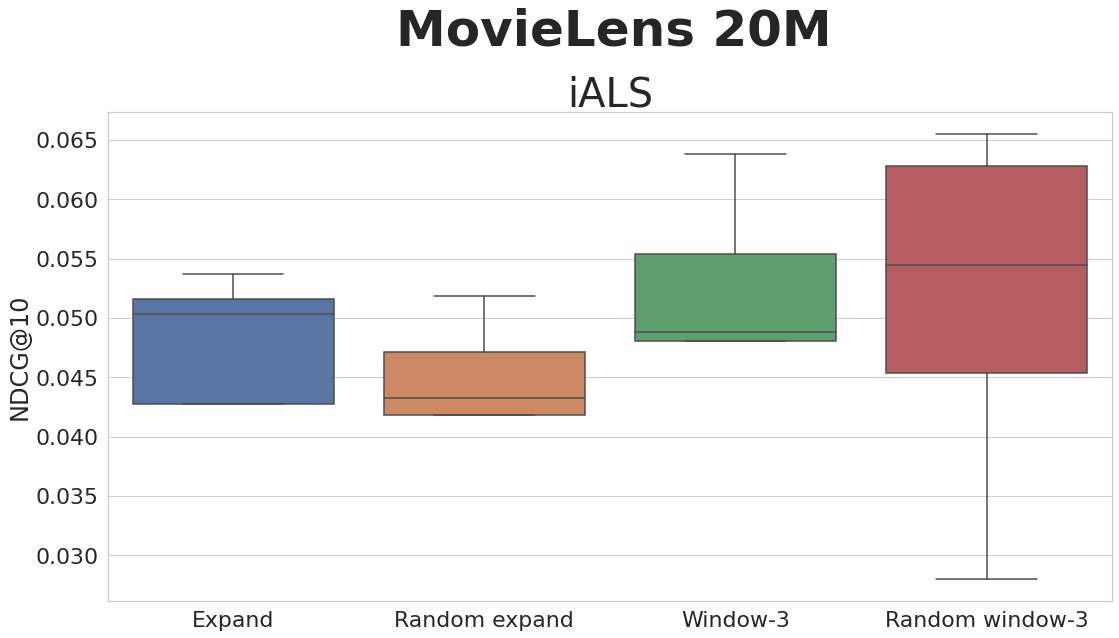}
    \end{subfigure}
    \begin{subfigure}[]
        \centering
        \includegraphics[width=0.32\textwidth]{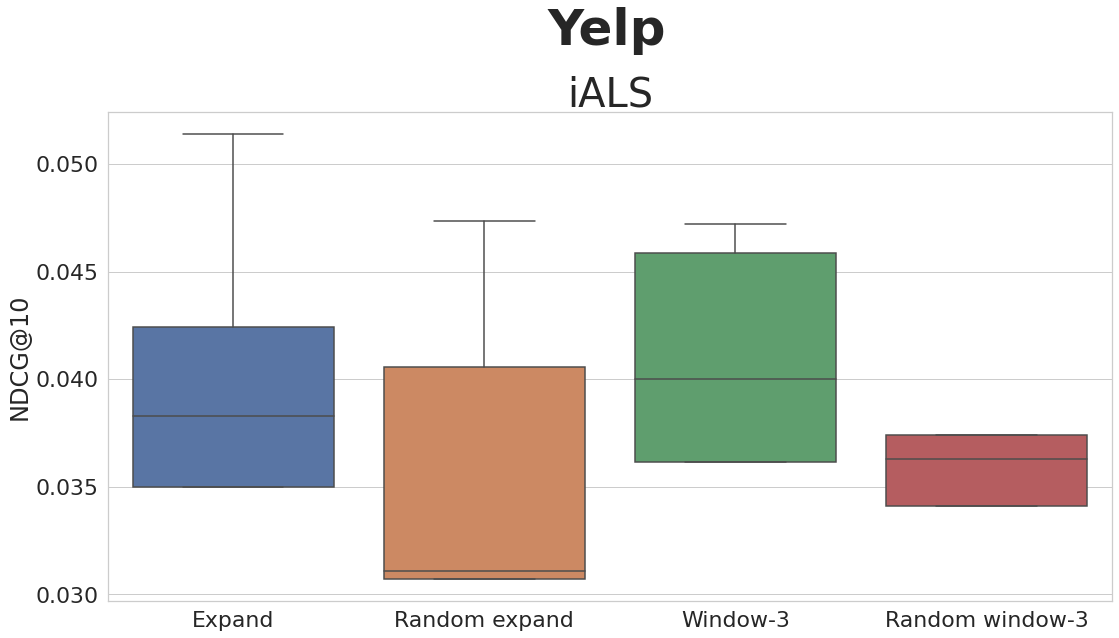}
    \end{subfigure}
    \begin{subfigure}[]
        \centering
        \includegraphics[width=0.32\textwidth]{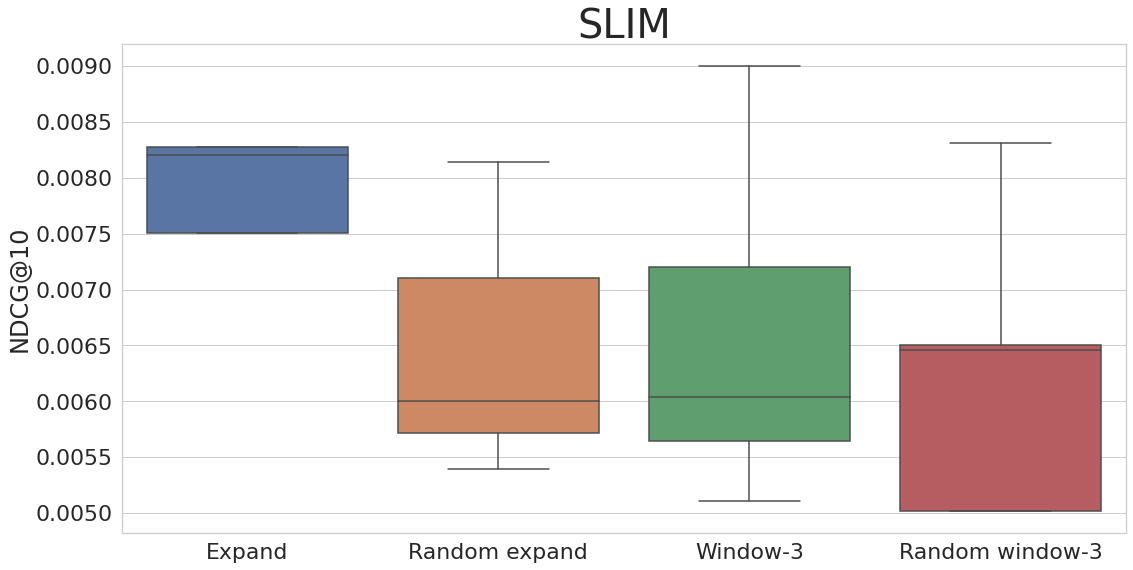}
    \end{subfigure}
    \begin{subfigure}[]
        \centering
        \includegraphics[width=0.32\textwidth]{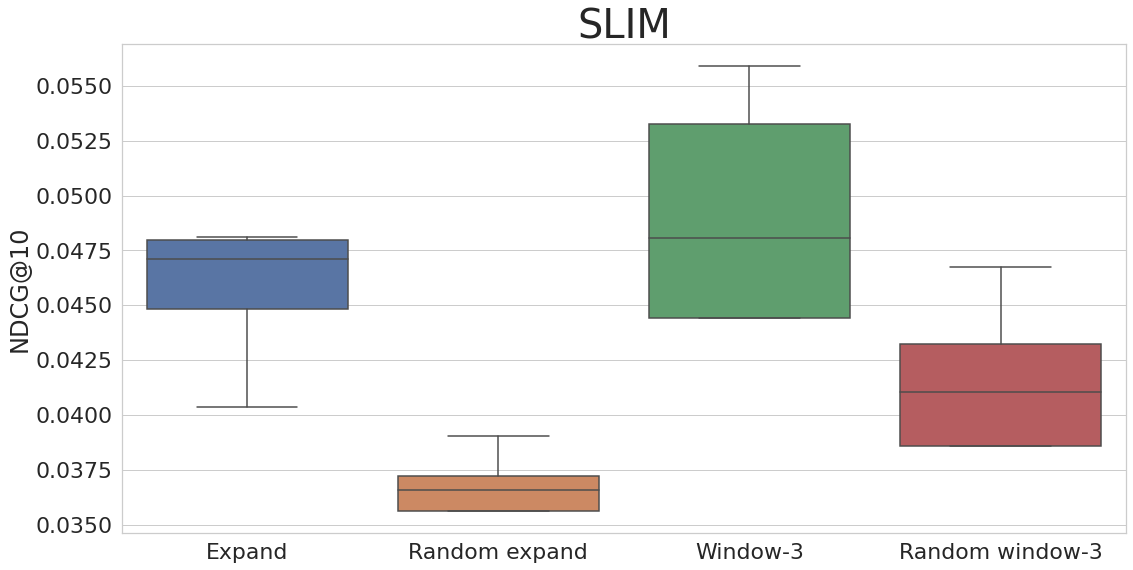}
    \end{subfigure}
    \begin{subfigure}[]
        \centering
        \includegraphics[width=0.32\textwidth]{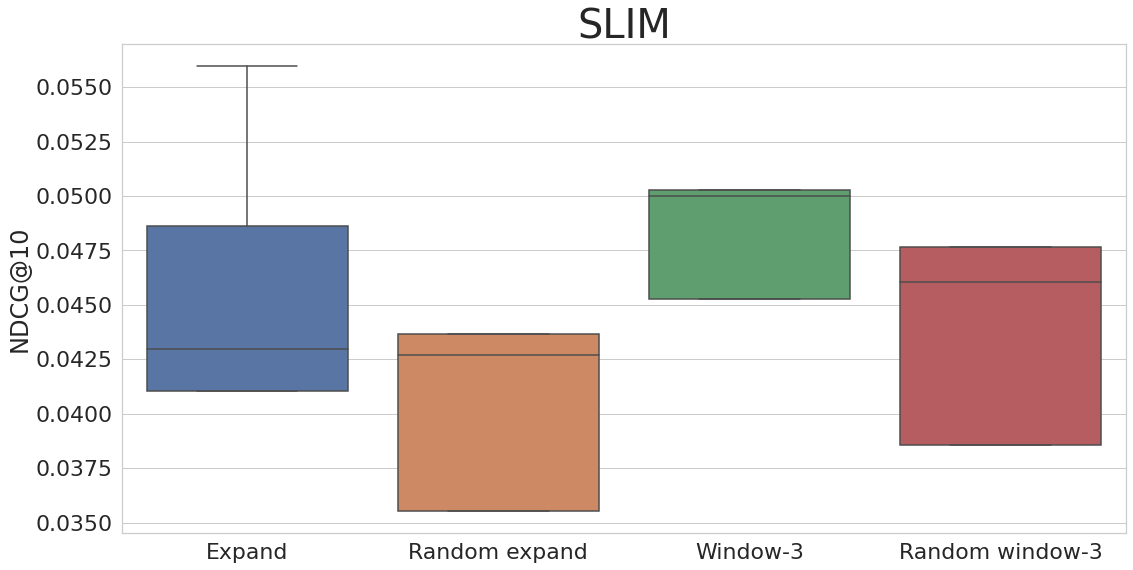}
    \end{subfigure}
    \begin{subfigure}[]
        \centering
        \includegraphics[width=0.32\textwidth]{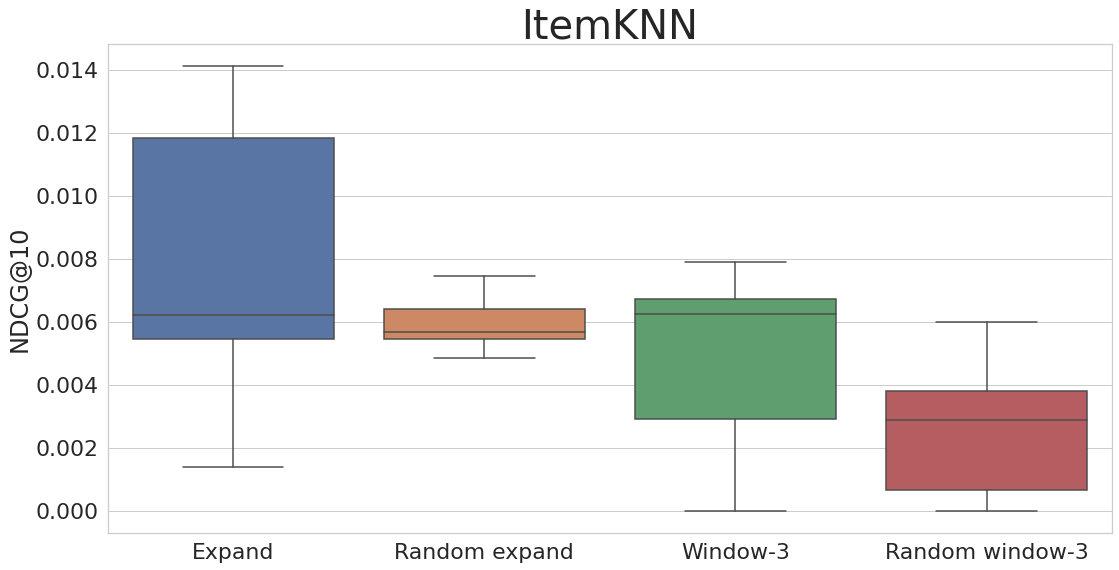}
    \end{subfigure}
    \begin{subfigure}[]
        \centering
        \includegraphics[width=0.32\textwidth]{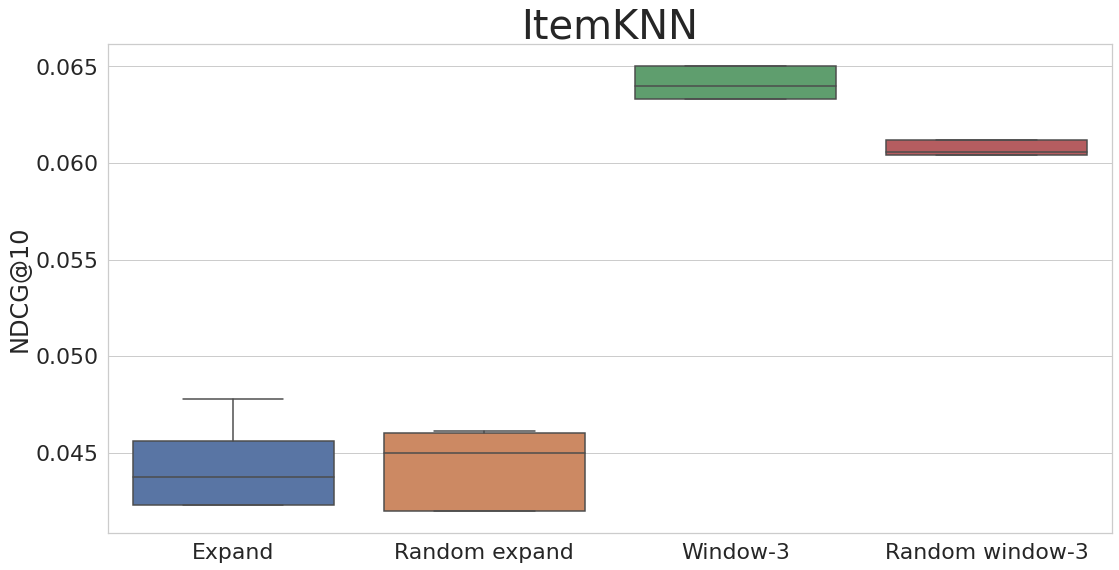}
    \end{subfigure}
    \begin{subfigure}[]
        \centering
        \includegraphics[width=0.32\textwidth]{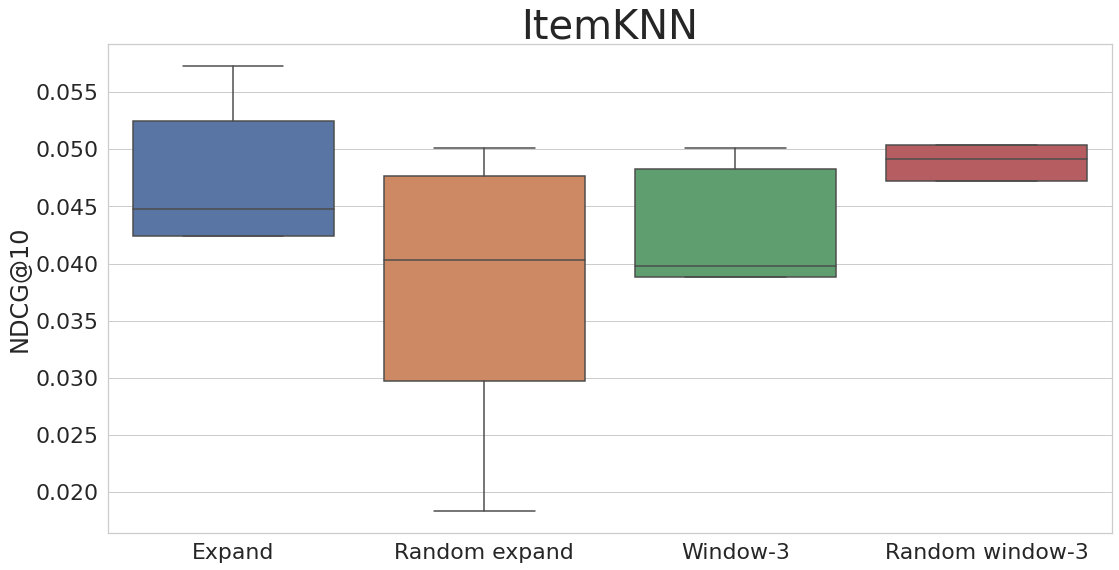}
    \end{subfigure}
    \begin{subfigure}[]
        \centering
        \includegraphics[width=0.32\textwidth]{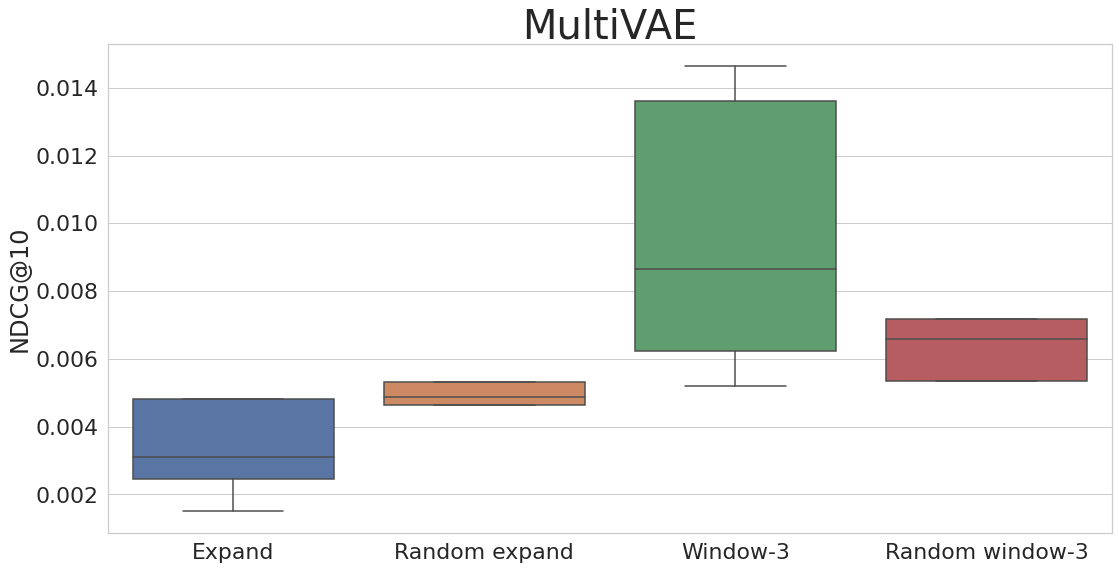}
    \end{subfigure}
    \begin{subfigure}[]
        \centering
        \includegraphics[width=0.32\textwidth]{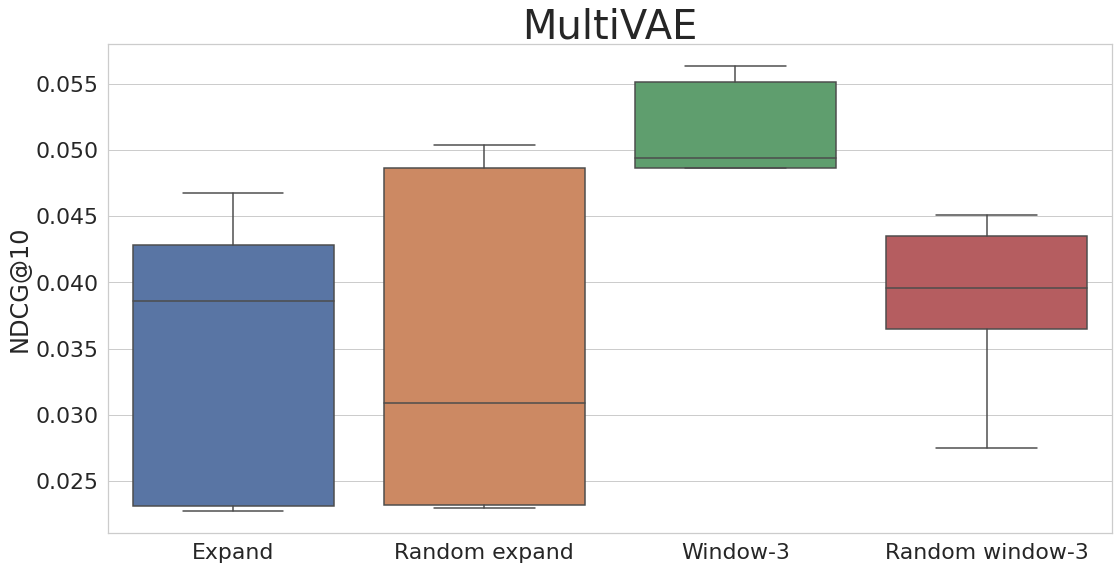}
    \end{subfigure}
    \begin{subfigure}[]
        \centering
        \includegraphics[width=0.32\textwidth]{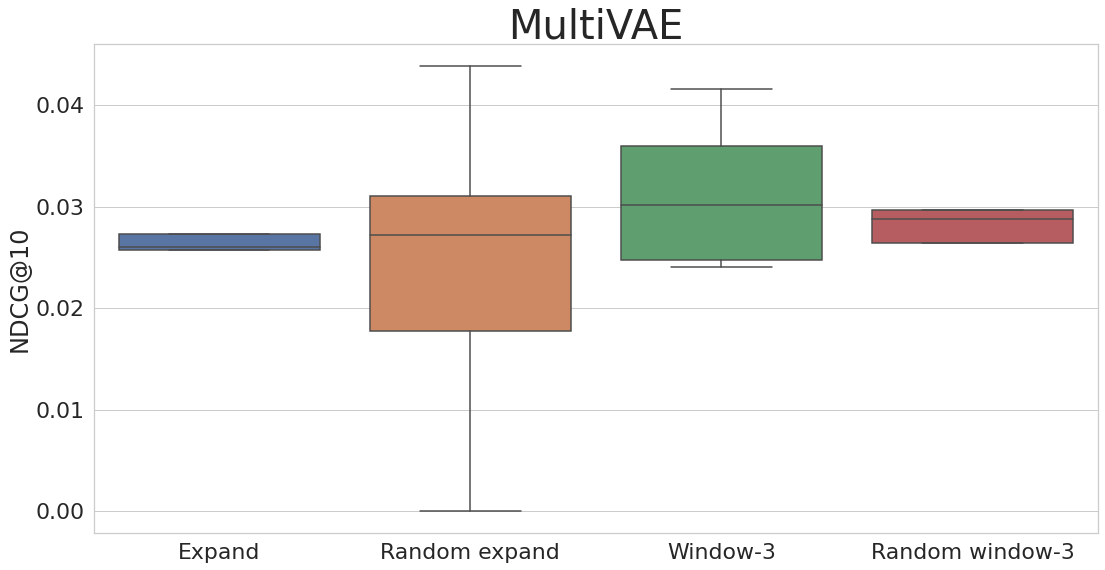}
    \end{subfigure}
    \caption{Method-centric performance comparison of CVTT temporal data-splitting strategies and their random-based versions.
Amazon Movies dataset is in the left column, MovieLens 20M in the middle, and Yelp in the right.
Each row corresponds to one of the evaluated methods, from top to bottom: iALS, SLIM, ItemKNN and MultiVAE.
X-axes correspond to the used data strategy and Y-axes to the performance metric ($NDCG@10$). 
}
    \label{fig:question1}
    \centering
\end{figure}

\subsection{Methods}
\label{sec:methods}

To determine the data-over-time effects on various different methods, we experiment with a set of 4 recommenders from the literature:
\begin{itemize}
    \item \textbf{IALS} \cite{als} is a matrix factorization-based (MF-based) model. 
    This model is designed to approximate any value in the interaction matrix by multiplying the user and item vectors in the hidden space. 
    \item \textbf{SLIM} \cite{ning2011slim} is a linear model that learns an item-item weight matrix to recommend new ones based on the previous history.
    \item \textbf{ItemKNN} \cite{itemknn} is an item-based k-nearest neighbors method, which utilizes similarities between previously purchased items. 
    Similar to \cite{dacrema2019we}, we used different similarity measures during our experiments: Jaccard coefficient, Cosine, Asymmetric Cosine, and Tversky similarity.
    \item \textbf{MultiVAE} \cite{liang2018variational} is a variational autoencoder approach (VAE-based) for a top-n recommendation task.
\end{itemize}

\subsection{Implementation Details}

Similar to previous studies \cite{dacrema2019we}, we search for the optimal parameters through Bayesian search using the implementation 
of Optuna \footnote{https://optuna.org}. 
For each $("train", "validation")$ subset, we iterate over 25 hypotheses.
During hyperparameter optimization, we use $NDCG@10$ metric for model selection.
We did not use any information from previous folds for the optimization, and the models were optimized and retrained from scratch at every fold.

\begin{figure}[t!]
    \centering
    \begin{subfigure}[]
        \centering
        \includegraphics[width=0.32\textwidth]{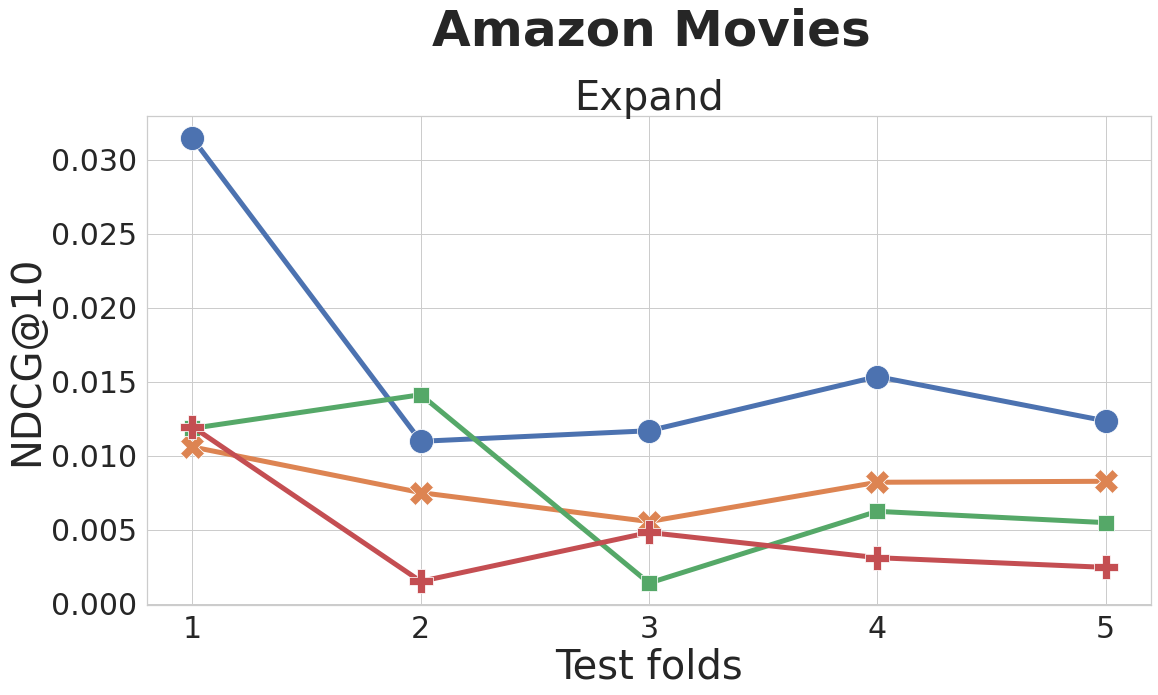}
    \end{subfigure}
    \begin{subfigure}[]
        \centering
        \includegraphics[width=0.32\textwidth]{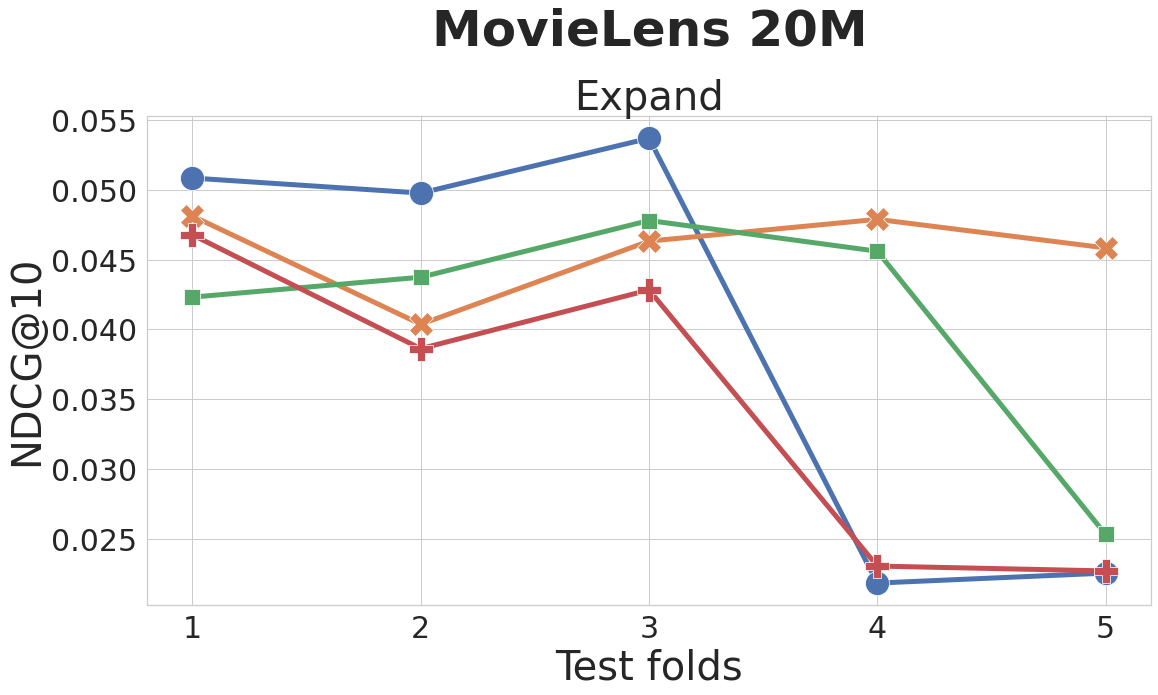}
    \end{subfigure}
    \begin{subfigure}[]
        \centering
        \includegraphics[width=0.32\textwidth]{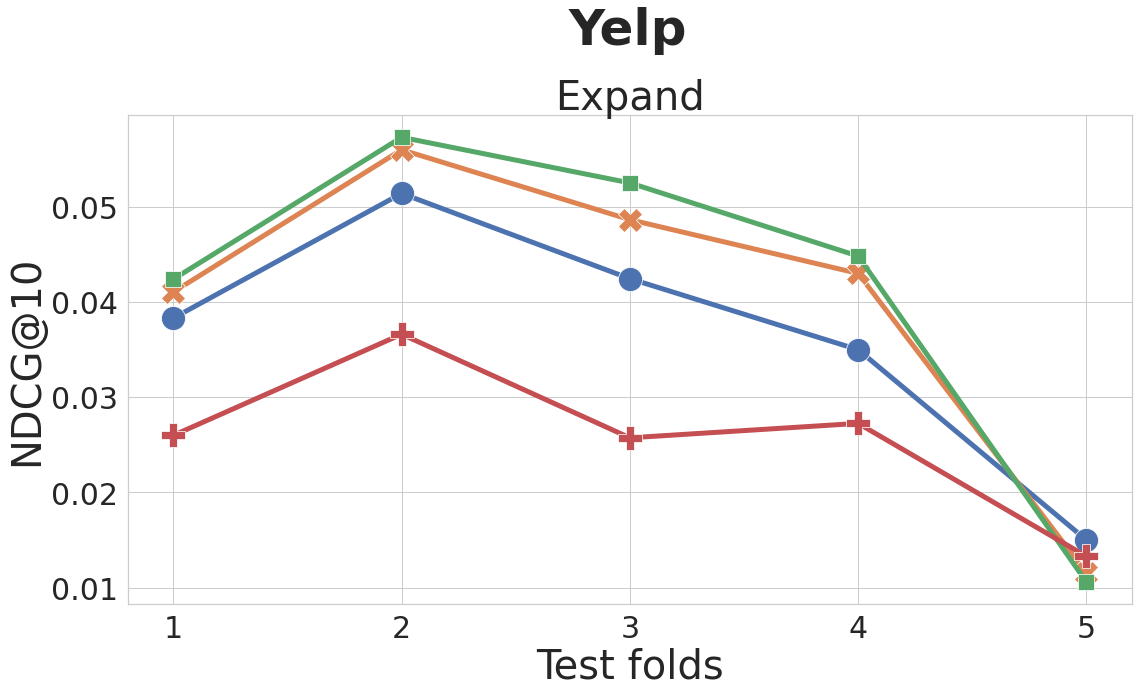}
    \end{subfigure}
    \begin{subfigure}[]
        \centering
        \includegraphics[width=0.32\textwidth]{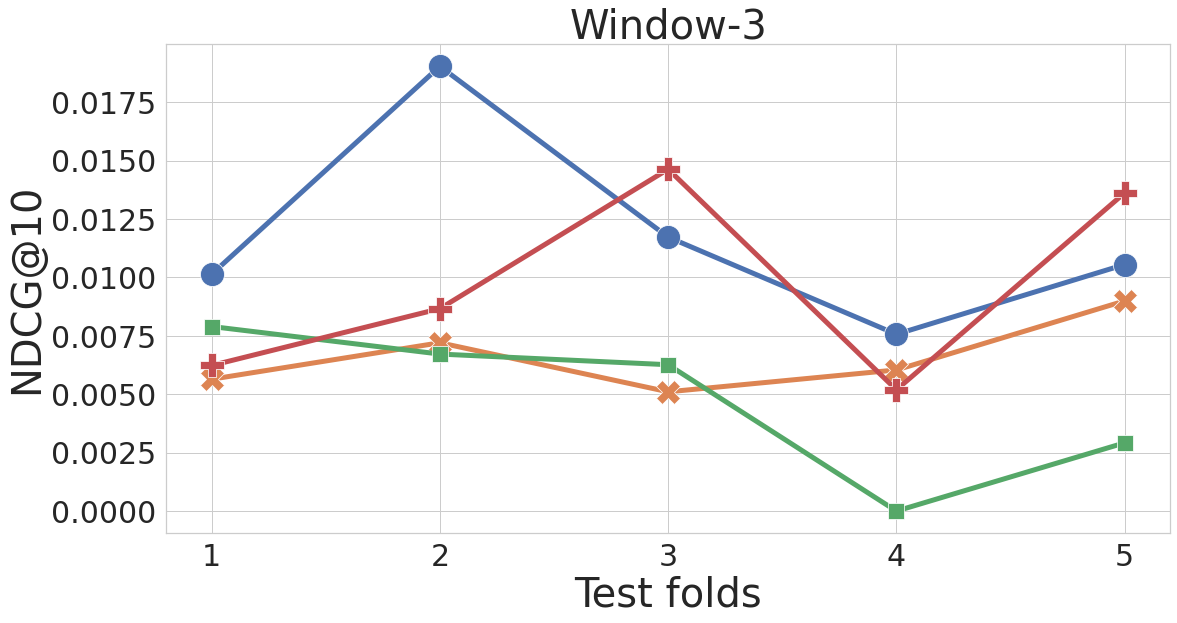}
    \end{subfigure}
    \begin{subfigure}[]
        \centering
        \includegraphics[width=0.32\textwidth]{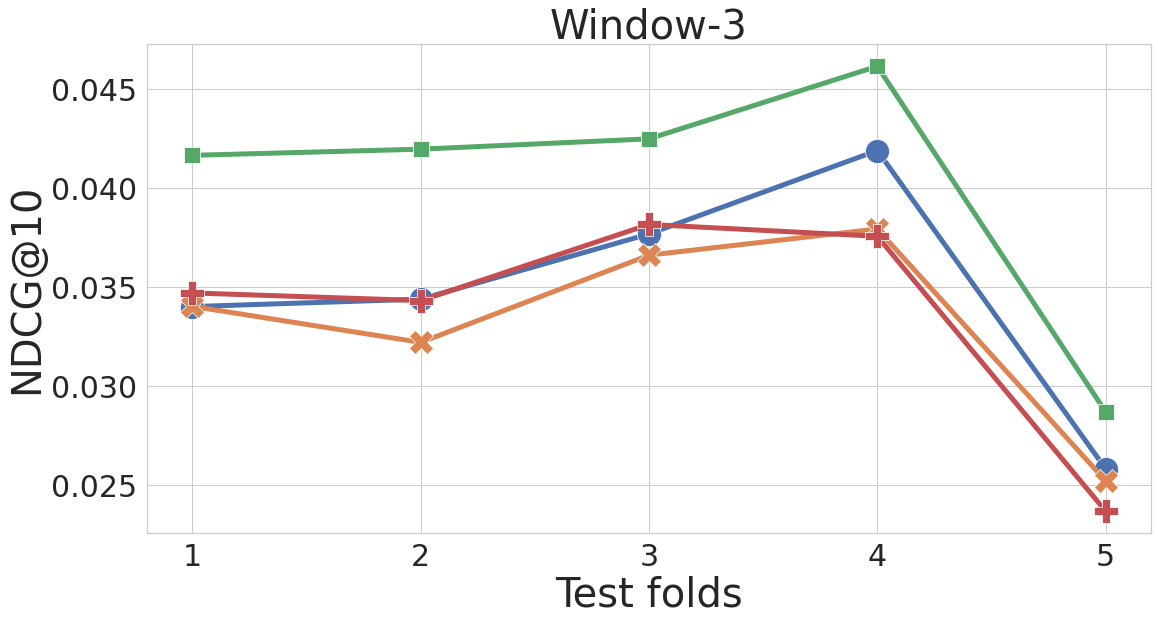}
    \end{subfigure}
    \begin{subfigure}[]
        \centering
        \includegraphics[width=0.32\textwidth]{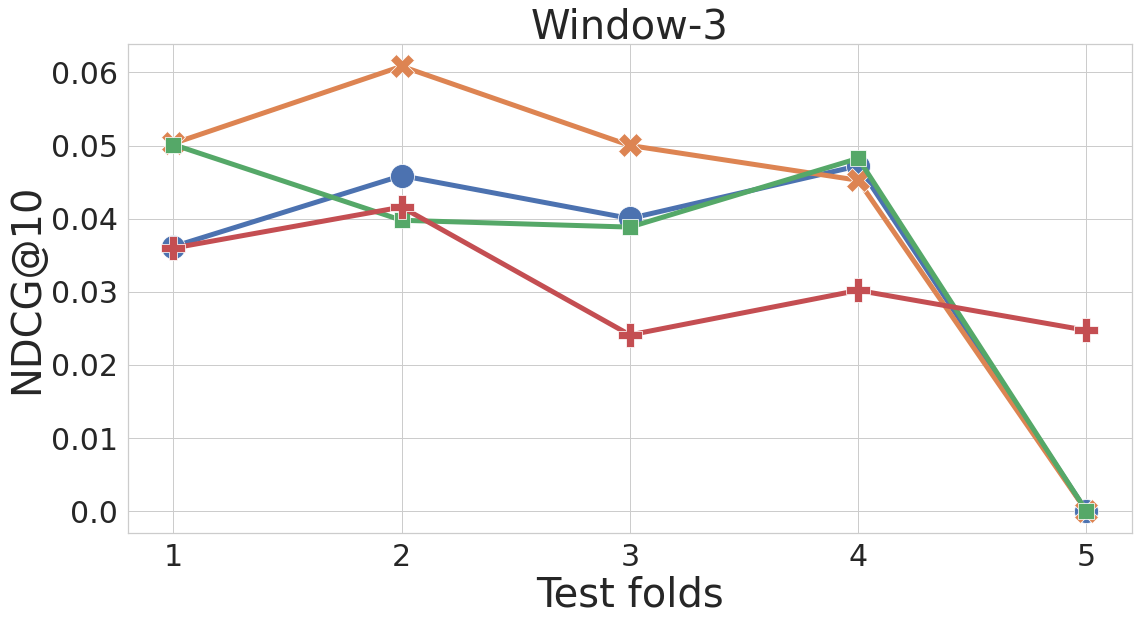}
    \end{subfigure}
    \begin{subfigure}
        \centering
        \includegraphics[width=0.5\textwidth]{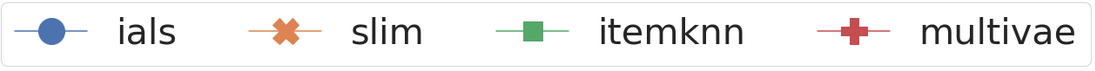}
    \end{subfigure}
    \caption{CVTT performance graphs. X-axes correspond to the measured test fold number and Y-axes to the performance metric ($NDCG@10$). (a, b, c) relates to expanding data strategy, while (d, e, f) correspond to window strategies with a window length of $3$.}
    \label{fig:question2}
    \centering
\end{figure}

\subsection{Comparing temporal and random split}
To verify the need for a global temporal split within CVTT, we compare it with a well-known random split, commonly used in the  community \cite{kaya2020ensuring, zhang2020content, zhou2020tafa}.
Following CVTT procedure (\autoref{fig:teaser}), we use different data splitting strategies over the "train" and "valid" parts, leaving the time-based evaluation during the "test" part.
The aggregated results over time are depicted in \autoref{fig:question1}.
We compare the Expand and Window-3 methods with their random-based versions in pairs. We note that in most pairs (18 of 24), performance on temporary-based partitions is better than on their random-based versions (i.e, iALS, SLIM, and ItemKNN on all datasets). For the MultiVAE model, Expand method loses to its random-based version, but on Window-3 it is slightly better on all datasets than the random-based version. 
We argue that such a performance gap indicated the need for broader usage of more realistic data-splitting approaches, such as a global temporal, rather than a random one.

\begin{figure}[t!]
    \centering
    \begin{subfigure}[]
        \centering
        \includegraphics[width=0.32\textwidth]{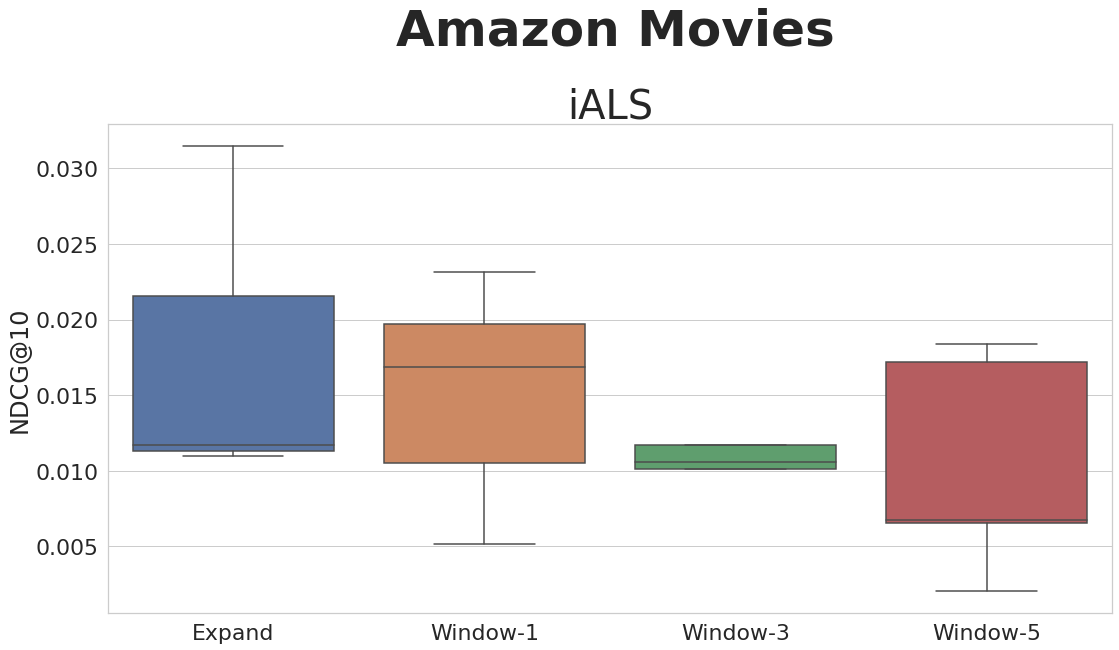}
    \end{subfigure}
    \begin{subfigure}[]
        \centering
        \includegraphics[width=0.32\textwidth]{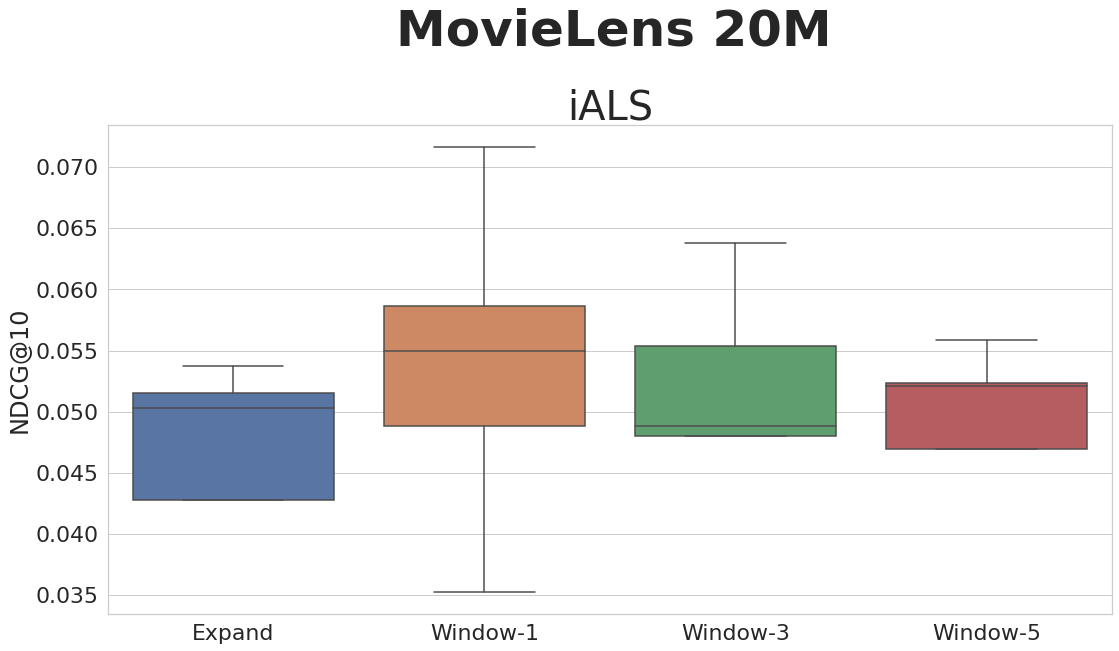}
    \end{subfigure}
    \begin{subfigure}[]
        \centering
        \includegraphics[width=0.32\textwidth]{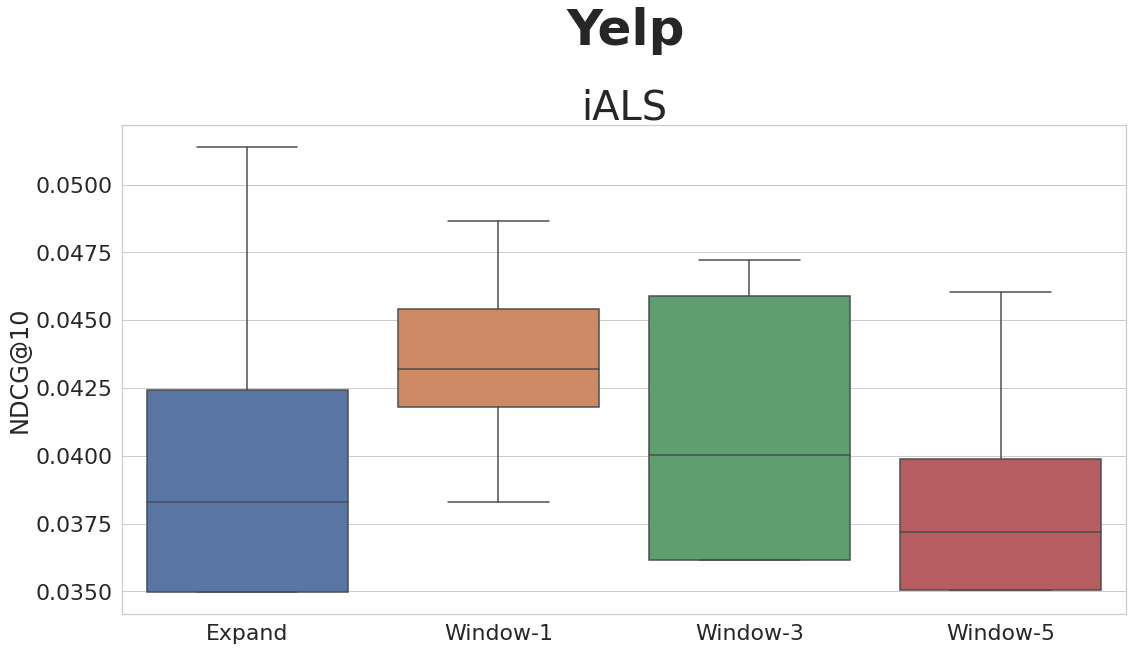}
    \end{subfigure}
    \begin{subfigure}[]
        \centering
        \includegraphics[width=0.32\textwidth]{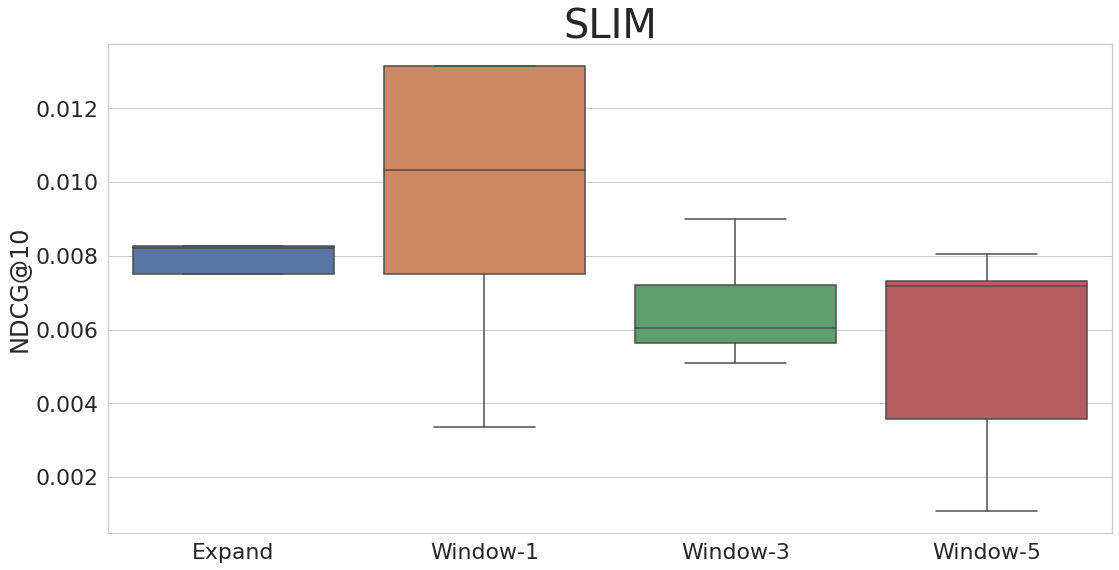}
    \end{subfigure}
    \begin{subfigure}[]
        \centering
        \includegraphics[width=0.32\textwidth]{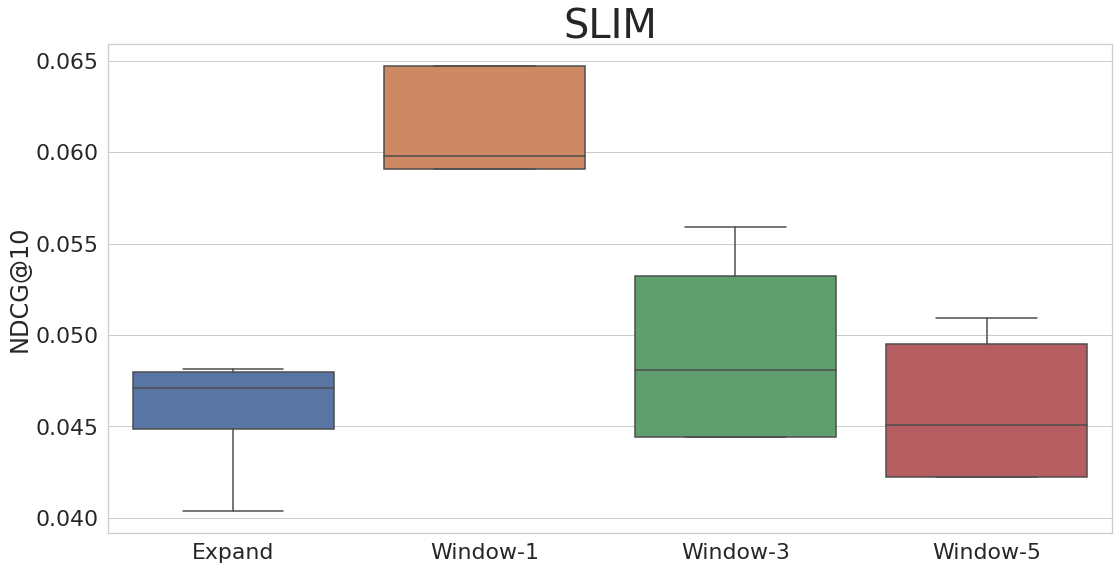}
    \end{subfigure}
    \begin{subfigure}[]
        \centering
        \includegraphics[width=0.32\textwidth]{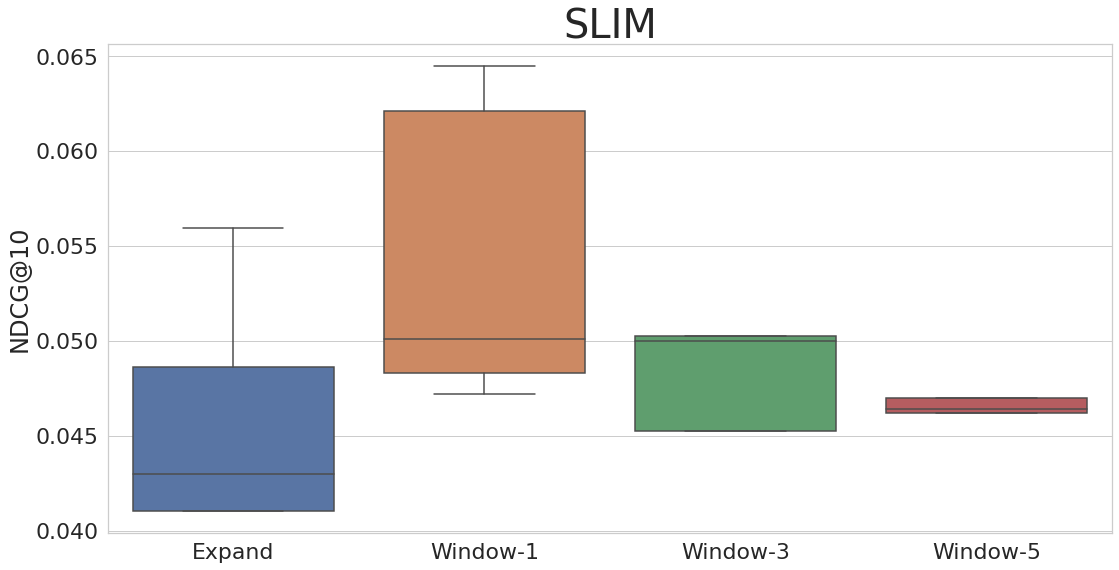}
    \end{subfigure}
    \begin{subfigure}[]
        \centering
        \includegraphics[width=0.32\textwidth]{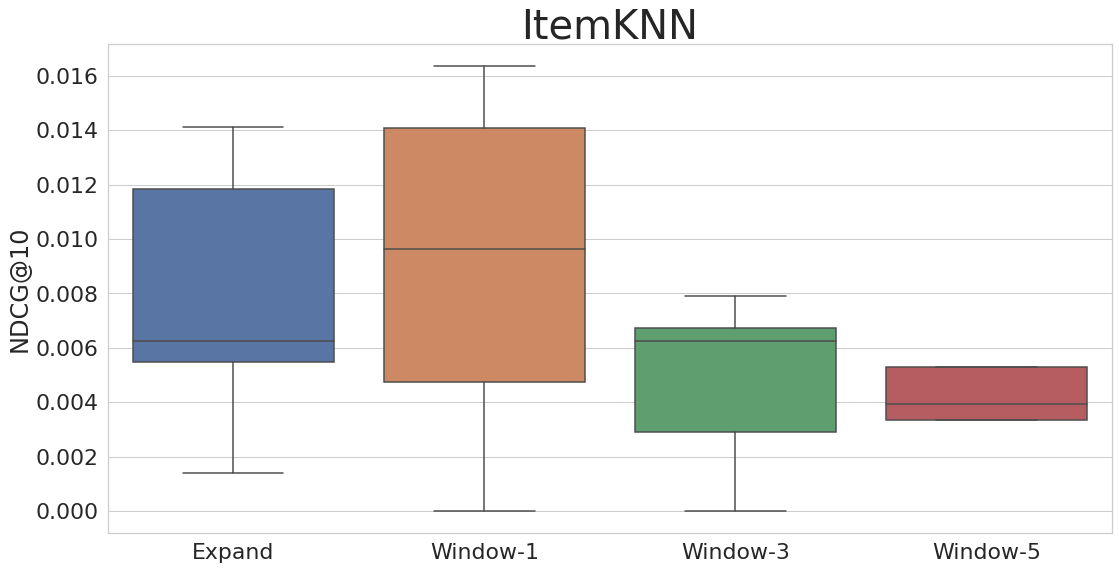}
    \end{subfigure}
    \begin{subfigure}[]
        \centering
        \includegraphics[width=0.32\textwidth]{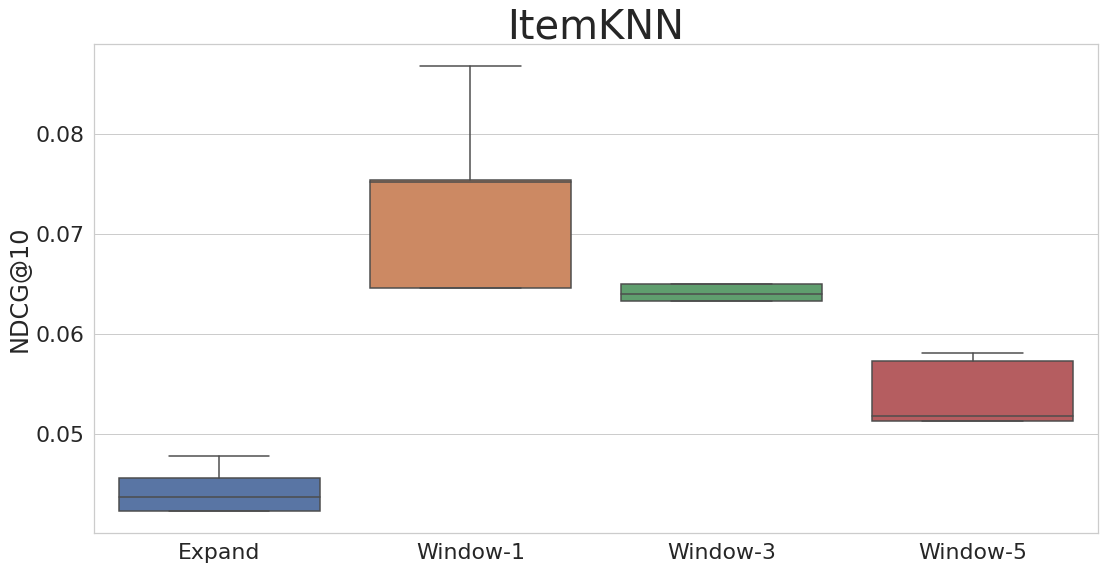}
    \end{subfigure}
    \begin{subfigure}[]
        \centering
        \includegraphics[width=0.32\textwidth]{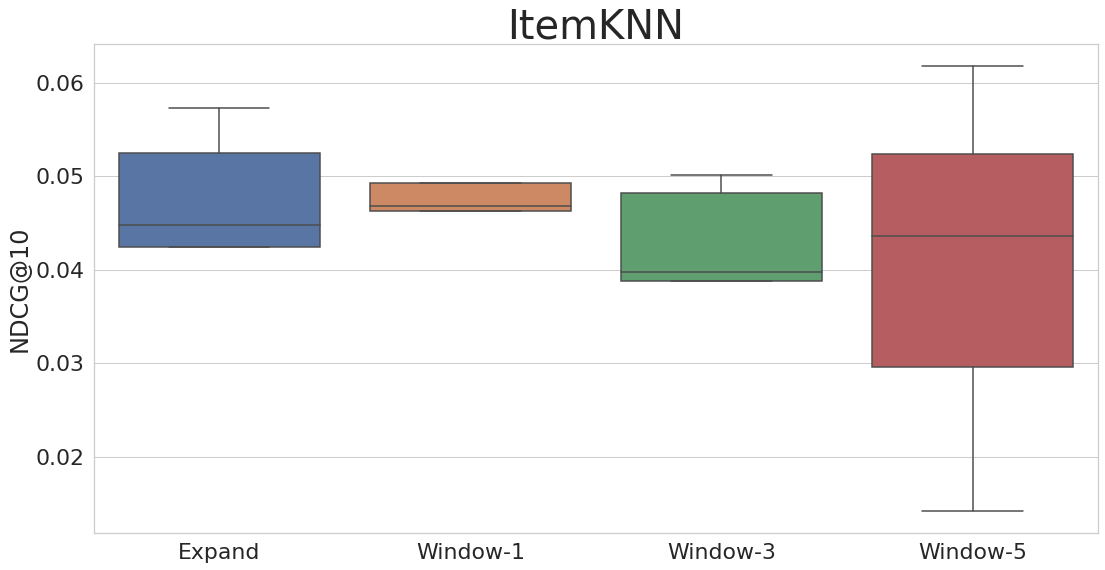}
    \end{subfigure}
    \begin{subfigure}[]
        \centering
        \includegraphics[width=0.32\textwidth]{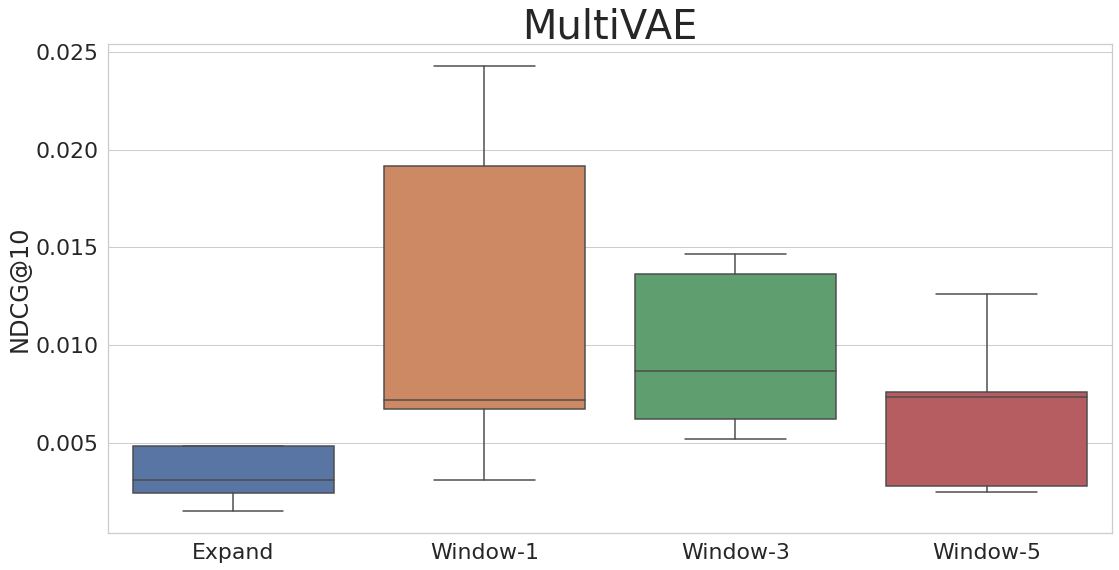}
    \end{subfigure}
    \begin{subfigure}[]
        \centering
        \includegraphics[width=0.32\textwidth]{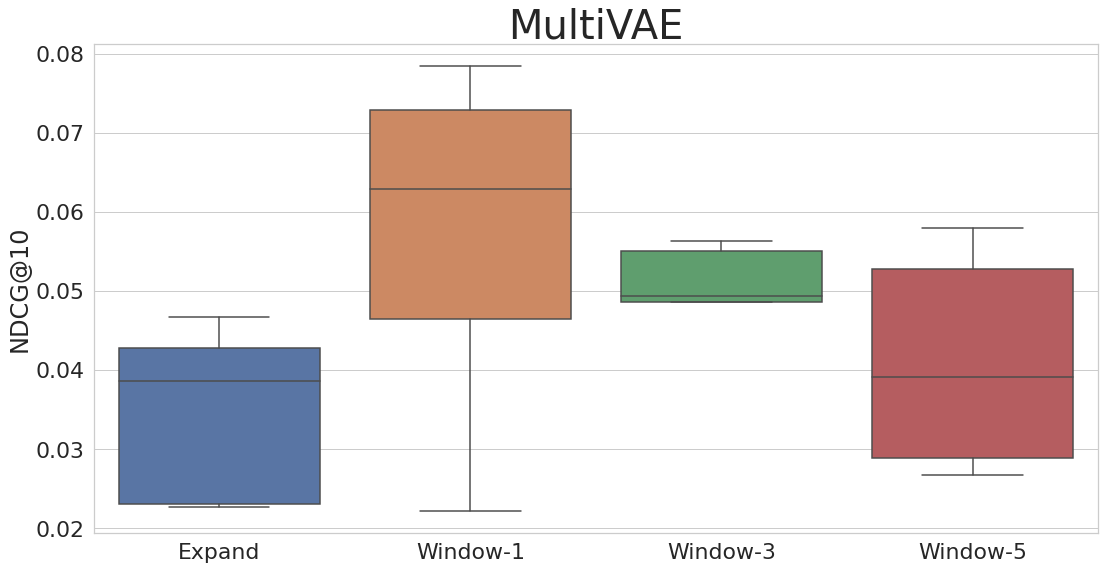}
    \end{subfigure}
    \begin{subfigure}[]
        \centering
        \includegraphics[width=0.32\textwidth]{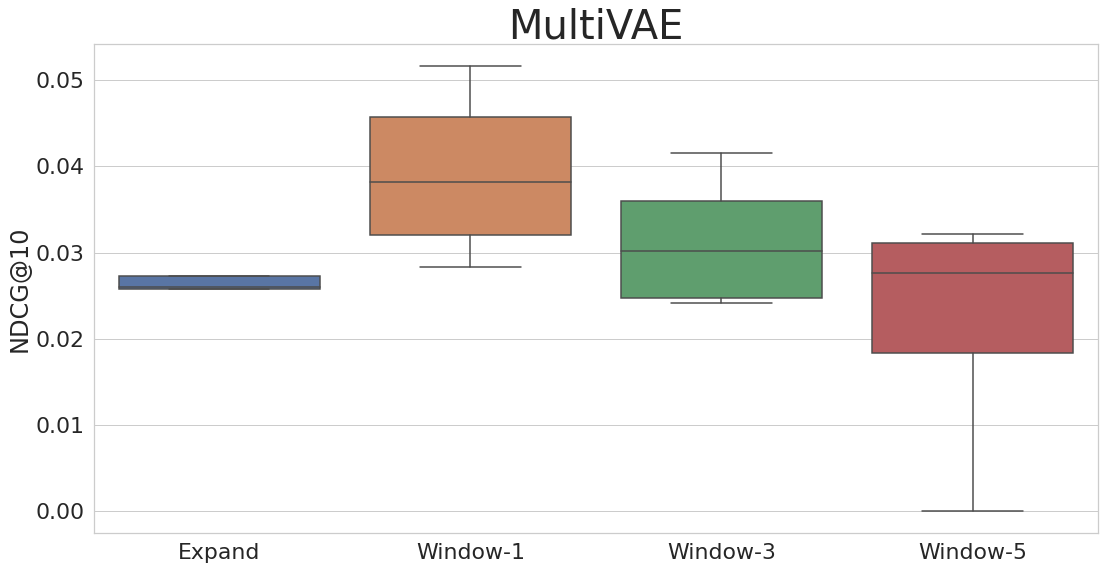}
    \end{subfigure}
    \caption{Method-centric CVTT performance comparison for different data strategies.
Amazon Movies dataset is in the left column, MovieLens 20M in the middle, and Yelp in the right.
Each row corresponds to one of the evaluated methods, from top to bottom: iALS, SLIM, ItemKNN and MultiVAE.
X-axes correspond to the used data strategy and Y-axes to the performance metric ($NDCG@10$). 
}
    \label{fig:question3}
    \centering
\end{figure}

\subsection{Evaluation over time}

In \autoref{fig:question2}, we contrast the performance of various methods over time following the CVTT evaluation protocol, expanding the training dataset at each fold.
Similar to previous studies \cite{44265d57d70f4725bf739c67800a5f77}, our analysis shows that the performance of evaluated approaches often changes over time.
Moreover, while performance remains roughly the same over time in most cases (i.e., SLIM on Amazon Movies), we observe that some data updates
can shift it significantly for some methods (i.e., iALS, SLIM, ItemKNN in (f)) while having no effect on others (i.e., MultiVAE in (f)). These findings open up a new set of questions concerning the effects of data flow on model performance.
Single-value evaluation could not catch such insights, leaving out these joint data-method effects. One possible way to underline them correctly is to evaluate methods continuously over time, which is the core idea behind CVTT.

\subsection{Effects of Data Preparation Strategies}

As discussed in Section 4, CVTT could work with $expand$ and $window$ strategies. While the  $expand$ one is commonly used in literature \cite{44265d57d70f4725bf739c67800a5f77}, we additionally explore a $window$-based strategy with different window lengths. The results for $expand$, $1$, $3$, and $5$ window lengths could be found in \autoref{fig:question3}. Several observations can be made based on these results.
First, in most cases for all datasets, the model performance decreases as the amount of training data used increases (i.e., $(e)$, $(f)$, $(h)$, $(j)$, $(k)$ ).
Secondly, the widely known $expand$ strategy achieves the best performance only in $(a)$, and $(i)$, suggesting that further research on data preparation approaches is required. 
A noteworthy implication of this analysis is that it is feasible to circumvent the requirement of furnishing a model with a comprehensive record of user interactions by adjusting the window length hyperparameter to optimize performance. In other words, utilizing a mere sufficient history amount that would cover current user preferences may prove to be sufficient.

\section{Conclusion}

In this study, we propose a novel approach by integrating recent developments in the areas of data partitioning, cross-validation, and offline evaluation of recommendation systems. We introduce CVTT, a straightforward method that utilizes temporal model performance for a more nuanced comparison. The proposed pipeline demonstrates the significance of temporal-based data partitioning, illustrates variations in recommendation system performance over time, and highlights the risk of suboptimal results when utilizing excessive data.


\bibliographystyle{ACM-Reference-Format}
\bibliography{main}

\clearpage
\appendix
\section{Appendix}
\clearpage
\begin{table}[!tb]
	\caption{\ricr{Overview of data splitting strategies reported in the literature, as well as the dataset(s) those papers use.}}
	\vspace{-3mm}
	\label{tab:models}
	\resizebox{1.0\textwidth}{!}{
		\begin{tabular}{cccccccp{1.6cm}}
			\toprule
			\multirow{2}{*}{\textbf{Model}} &
			\multirow{2}{*}{\textbf{Random Split}}& 
			\multirow{2}{*}{\textbf{User Split}} &
			\multirow{2}{*}{\textbf{Leave One Out}} & 
			\multicolumn{2}{c}{\textbf{Temporal Split}} &  
			\multirow{2}{*}{\textbf{Used Datasets}} \\
			\cmidrule{5-6}
			&   &  &  &User-based & Global \\
			\midrule
			\midrule
			BPR~\cite{rendle2009bpr} (2009) & $\surd$ &$\times$ &  $\times$ & $\times$ & $\times$  & N\\
			AVG~\cite{baltrunas2010group} (2010) & $\surd$ & $\surd$ & $\times$ & $\times$ & $\times$  & M100\\
			FPMC~\cite{rendle2010factorizing} (2010) &$\times$ & $\times$ & $\surd$ & $\times$ & $\times$ & -\\
			PureSVD~\cite{cremonesi2010performance} (2010) & $\surd$ & $\times$ & $\times$ & $\times$ & $\surd$ & M1, N\\
            SLIM~\cite{ning2011slim} (2011) & $\surd$ & $\times$ & $\times$ & $\times$ & $\times$ & M10, N, B\\
            AutoRec~\cite{sedhain2015autorec} (2015) & $\surd$ & $\times$ & $\times$ & $\times$ & $\times$ & M1, M10, N\\
            CDAE~\cite{wu2016collaborative} (2016) & $\surd$ & $\times$ & $\times$ & $\times$ & $\times$ & M10, Y, N\\
            VBPR~\cite{he2016vbpr} (2016) & $\surd$ & $\times$ & $\times$ & $\times$ & $\times$ & A\\
            CML~\cite{hsieh2017collaborative} (2017) & $\surd$ & $\times$ & $\times$ & $\times$ & $\times$ & M2, B\\
			NeuMF~\cite{he2017neural} (2017) & $\times$ &$\times$ &  $\surd$ & $\times$ & $\times$ & M1, P\\
			CVAE~\cite{li2017collaborative} (2017) & $\surd$ & $\times$ & $\times$ & $\times$ & $\times$ & -\\
			GreedyLM~\cite{xiao2017fairness} (2017) & $\surd$ & $\surd$ & $\times$ & $\times$ & $\times$ & M1\\
			VAECF~\cite{liang2018variational} (2018) & $\times$  & $\surd$ & $\times$ &  $\surd$ & $\times$  & M2, N\\
			Triple2vec~\cite{wan2018representing} (2018) & $\times$ & $\times$ & $\surd$ &  $\times$ & $\times$ & I, D\\
			SASRec~\cite{kang2018self} (2018) & $\times$ & $\times$ & $\surd$  & $\times$ & $\times$  & A, M1\\
			TARMF~\cite{lu2018coevolutionary} (2018) & $\surd$ & $\times$ & $\times$ & $\times$ & $\times$ & Y, A\\
            XPO~\cite{sacharidis2019topn} (2019)& $\surd$ & $\surd$ & $\times$ & $\times$ & $\times$ & M1\\
            CTRec ~\cite{bai2019ctrec} (2019) & $\times$ & $\times$ & $\surd$ & $\surd$  & $\times$ & T, A\\
			SVAE ~\cite{sachdeva2019sequential} (2019) & $\times$ & $\surd$ & $\times$ & $\surd$ & $\times$  & M1, N\\
			BERT4Rec ~\cite{sun2019bert4rec}(2019) &  $\times$ & $\times$ & $\surd$  & $\times$  & $\times$ & A, M1, M2\\
			VBCAR~\cite{meng2019variational} (2019) &$\times$ & $\times$ & $\times$ & $\times$ & $\surd$  & I\\
            Set2Set~\cite{hu2019sets2sets} (2019) &$\times$ & $\times$ & $\times$ & $\surd$ &  $\times$ & T, D\\
			DCRL~\cite{xiao2019dynamic} (2019) & $\times$ &$\times$ &$\times$ & $\times$ & $\surd$ & M2, G\\
			MARank~\cite{yu2019multi} (2019) & $\times$ & $\times$ & $\surd$ & $\times$ & $\times$ & Y, A\\
			GATE~\cite{ma2019gated} (2019) & $\surd$ & $\times$ & $\times$ & $\times$ & $\times$ & M2, B\\
			TiSASRec~\cite{li2020time} (2020) & $\times$ &$\times$ & $\surd$   & $\times$ & $\times$ & M1, A\\
            JSR~\cite{zamani2020learning} (2020) & $\times$ & $\times$ & $\surd$  & $\times$ & $\times$ & M2, A\\
            TAFA~\cite{zhou2020tafa} (2020) & $\surd$ & $\times$ & $\times$  & $\times$ & $\times$ & Y, A\\
            DICER~\cite{zhang2020content} (2020) & $\surd$ & $\times$ & $\times$ & $\times$ & $\times$ & A\\
            GFAR~\cite{kaya2020ensuring} (2020) & $\surd$ & $\surd$ & $\times$  & $\times$ & $\times$ & M1, K\\
             \bottomrule
			\multicolumn{7}{c}{M1: Movielens-1M, M10: Movielens-10M, M2: Movielens-20M, M100:Movielens-100K }\\
			\multicolumn{7}{c}{G: Gowalla, I: Instacart, A: Amazon, Y: Yelp, P: Pinterest}\\
			\multicolumn{7}{c}{B: Book-X, D: Dunnhumby, T: Tafeng, K: KGRec-music, N: Netflix}
		\end{tabular}
	}
	\vspace{-3mm}
\end{table}

\begin{table*}[!h]
	\centering
	\caption{Hyperparameters search space.}
	\vskip 0.15in
	\scalebox{0.99}{\begin{tabular}{cc}
		\toprule
		\textbf{Algorithms} & \textbf{Search Space}\\
		
		\midrule
		\textbf{SLIM} & \begin{tabular}[c]{@{}c@{}}
		l1 ratio $\in [1e-5, 1.0]$\\
		$\alpha \in [1e-3, 1.0]$\\
		positive only $\in$ \{True, False\}\\
		top k $\in [5, 800]$\\
		    \end{tabular}
		\\
		\midrule
		\textbf{iALS} & \begin{tabular}[c]{@{}c@{}}
		confidence scaling $\in$ \{True, False\}\\
		number of factors $\in [1, 200]$\\
		$\alpha \in [1e-3, 50.0]$\\
		$\epsilon \in [1e-3, 10.0]$\\
		regularization $\in [1e-5, 1e-2]$\\
		    \end{tabular}
		\\
		\midrule
		\textbf{ItemKNN} & \begin{tabular}[c]{@{}c@{}}
		top k $\in [1, 200]$\\
		shrink $\in [0, 600]$\\
		similarity $\in$ \{cosine, jaccard, asymmetric, dice, tversky\}\\
		    \end{tabular}
		\\
        \midrule
		\textbf{MultiVAE} & \begin{tabular}[c]{@{}c@{}}
		total anneal steps $\in [10000, 500000]$\\
		hidden layer size $\in [32, 600]$\\
		latent dimension $\in [32, 600]$\\
        encoder layers $\in [1, 4]$\\
        dropout probabilities $\in [0.05, 0.5]$\\
		$\beta \in [0.2, 1.0]$\\
		    \end{tabular}
		\\
        \bottomrule
	\end{tabular}}
	\label{tab:hparams}
\end{table*}

\end{document}